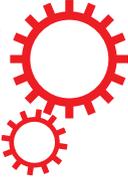

# OPEN

# Interactive Outlining of Pancreatic Cancer Liver Metastases in Ultrasound Images

Jan Egger[1,2], Dieter Schmalstieg[1], Xiaojun Chen[3], Wolfram G. Zoller[4] & Alexander Hann[5]

Ultrasound (US) is the most commonly used liver imaging modality worldwide. Due to its low cost, it is increasingly used in the follow-up of cancer patients with metastases localized in the liver. In this contribution, we present the results of an interactive segmentation approach for liver metastases in US acquisitions. A (semi-) automatic segmentation is still very challenging because of the low image quality and the low contrast between the metastasis and the surrounding liver tissue. Thus, the state of the art in clinical practice is still manual measurement and outlining of the metastases in the US images. We tackle the problem by providing an interactive segmentation approach providing real-time feedback of the segmentation results. The approach has been evaluated with typical US acquisitions from the clinical routine, and the datasets consisted of pancreatic cancer metastases. Even for difficult cases, satisfying segmentations results could be achieved because of the interactive real-time behavior of the approach. In total, 40 clinical images have been evaluated with our method by comparing the results against manual ground truth segmentations. This evaluation yielded to an average Dice Score of 85% and an average Hausdorff Distance of 13 pixels.

Compared to computed tomography (CT) or magnetic resonance imaging (MRI), ultrasound (US) is a more easily accessible and less expensive imaging procedure, and, for the liver, the most commonly used imaging modality worldwide. The preference for US is owed to the multiple access points of the liver for the US examination. The intercostal and subcostal route allow for the examination of the whole organ in most of the patients. Hence, the usage of US is also one of the first procedures for evaluating liver metastases, when a patient is diagnosed with cancer. After a subsequent initial staging using CT of the thorax and the abdomen, US is often used to evaluate for treatment response of cancer patients with metastases solely localized in the liver. In case of pancreatic cancer, additional staging of the primary tumor can be done with endoscopic ultrasound. Due to the good accuracy of US in diagnosis and follow-up of liver metastases compared to CT or MRI[1], the current ESMO – ESDO clinical practice guideline recommend US for the response evaluation in the palliative setting of pancreatic cancer patients[2]. Additionally to the appearance of a new metastases during follow-up, the change in size of preexisting metastases plays an important role in the evaluation of treatment response. Nevertheless, the liver metastases appearance in US acquisitions is highly variable[3]. A schematic overview of the different echo-patterns is presented in Fig. 1: Compared to the surrounding liver tissue, metastases can appear, for example, hyperechoic/brighter (A) or hypo-echoic/darker (C) in B-mode. However, isoechoic (B) masses can have a very similar echo-pattern compared to the surrounding liver tissue and can be hard to detect. Consequently, the size determination of isoechoic masses is also very challenging. A darker rim around the liver lesion, called hypoechoic halo, can often been seen around hyper (D) or isoechoic (E) metastases. All these different liver lesions appearances (A)-(E) plus additional morphologic changes over time contribute to a strong observer dependence for US data[4]. A solution to reduce some of the issues with the poor inter-observer agreement of US images could be an automatic tool to measure the size of a metastasis with different echogenic patterns. Thus, making a first step in a reliably evaluation of liver masses growth over time. Hence, in the first part of this study, we evaluated and adjusted an automatic segmentation method by using different echo-patterns of liver metastases. The groups included hyper-, hypoechoic and nearly isoechoic lesions. In a second step, we used the settings adjusted for the different echo pattern to evaluate the accuracy of the method using an additional set of 40 pancreatic cancer liver metastases.

[1]Institute for Computer Graphics and Vision, Graz University of Technology, Inffeldgasse 16, 8010, Graz, Austria. [2]BioTechMed, Krenngasse 37/1, 8010, Graz, Austria. [3]Shanghai Jiao Tong University, School of Mechanical Engineering, 800 Dong Chuan Road, Shanghai, 200240, China. [4]Department of Internal Medicine and Gastroenterology, Katharinenhospital, Stuttgart, Germany. [5]Department of Internal Medicine I, Ulm University, Ulm, Germany. Correspondence and requests for materials should be addressed to J.E. (email: egger@tugraz.at) or X.C. (email: xiaojunchen@163.com) or A.H. (email: Alexander.Hann@uniklinik-ulm.de)

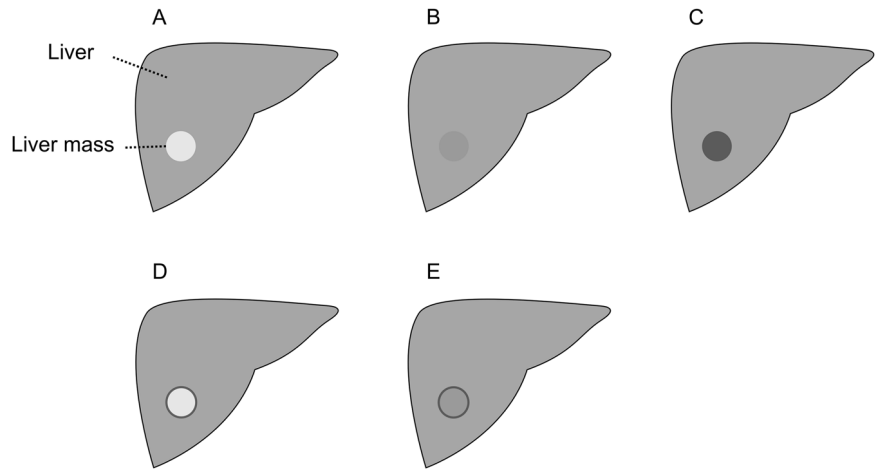

**Figure 1.** Different echo pattern of homogenous liver masses in ultrasound B-mode presented in a schematic overview. Relative to the liver echopattern, liver masses can appear hyperechoic (brighter) (**A**), isoechoic (similar) (**B**) or hypoechoic (darker) (**C**). Iso- and hyperechoic liver masses can present with a hypoechoic halo (rim) (**D,E**). Note: figure adapted from[26].

Other researchers working in the specific area of liver tumor segmentation in US images are Boen[5], Kumar et al.[6], Yoshida et al.[7], and Poonguzhali and Ravindran[8]. In contrast, Hiransakolwong et al.[9], Hao et al.[10] and Wang and Li[11] focus more on general segmentation of US images of the liver, which could also be used for tumor segmentation. Boen[5] introduce an approach called seeded region growing (SRG) that isolates regions of interest for subsequent processing. The SRG method involves selecting seed points (manually or automatically) as the starting locations for the segmentation process. For this paper, both methods of seed selection (manual/automatic) were tested and compared. The overall segmentation system was evaluated on simulated objects and B-mode ultrasound images of the liver with suspected lesions. A fully automatic segmentation method that uses statistical features to help distinguishing between normal and ultrasonic tumor liver images, has been presented by Kumar et al.[6]. Their approach uses a peak and valley method to filter the noise in the US images. In addition, the image is smoothed, and, in a second stage, the filter is adopted to further filter the noise and to improve the quality of the image. Yoshida et al.[7] introduce an approach for the segmentation of low-contrast objects embedded in noisy images. They used it to segment liver tumors in B-scan ultrasound images with hypoechoic rims. In a first step, the B-scan image is processed by a median filter to remove speckle noise, followed by obtaining several one-dimensional profiles along multiple radial directions, which pass through the manually identified center of the region of a tumor. Afterwards, these profiles are processed by Sombrero's continuous wavelets to yield scalograms over a range of scales. The modulus maxima lines are utilized for identifying candidate points on the boundary of the tumor. Finally, the detected boundary points are fitted by an ellipse and are used as an initial configuration of a wavelet snake, which is deformed to find a more accurate boundary of the tumor. An automatic method for the segmentation of masses in ultrasound images using a region growing algorithm has been proposed by Poonguzhali and Ravindran[8]. They automatically select a seed point for the region growing from the abnormal region based on textural features, such as co-occurrence features, and run length features. Additionally, the threshold to control the region growing process is also automatically selected, and a gradient magnitude based region growing algorithm is adopted. This segmentation approach was tested on different abdominal masses such as cyst and liver tumors. We also want to point the interested reader to a survey on liver CT image segmentation methods[12] and a review about the segmentation of the liver in CT and MRI images[13]. Finally, we want to mention that there exist several works on ultrasonography in other organs, like breast nodules[14–17], which could be adapted in the future to segment also liver metastases.

Previous approaches that use graph-based segmentation methods for US images are Zhang et al.[18], Zhou et al.[19] and Huang et al.[20, 21]. Zhang et al.[18] propose a fully automatic system to detect and segment breast tumors in 2D ultrasound images. For the tumor segmentation, they propose a discriminative graph cut approach, where both the data fidelity and compatibility functions are learned discriminatively. Zhou et al.[19] propose a new method for semi-automatic tumor segmentation on breast ultrasound (BUS) images using Gaussian filtering, histogram equalization, mean shift, and graph cuts. In doing so, the object and background seeds for the graph cuts were automatically generated on the pre-filtered images. Using these seeds, the images were afterwards segmented by the graph cut into a binary image containing the object and background. Finally, Huang et al.[20, 21] introduce a graph-based segmentation method and its optimization for breast tumors in ultrasound images. The method constructs a graph using improved neighborhood models. Additionally, taking advantages of local statistics, a pair-wise region comparison predicate that was insensitive to noises was proposed to determine the mergence of any two of adjacent subregions.

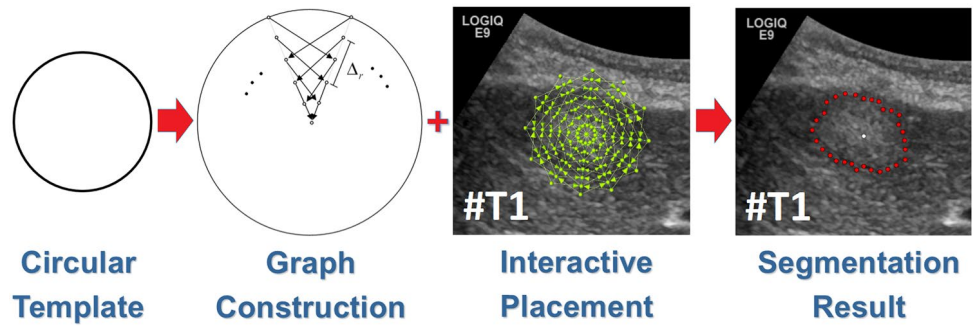

**Figure 2.** Interactive segmentation workflow. Left image: a circular template is used for the underlying graph. Second image from the left: based on the underlying circular template, the graph is constructed. Therefore, rays are sent out radially from the center of the circle template and along these rays the graphs' nodes are sampled. Afterwards, intra- and inter-edges are established between these nodes (note: the inter-edges depend on a delta value $\Delta_r$). Third image from the left: depending on the interactive placement of the mouse cursor by the user (which is also the center of the circle template/graph), the complete graph (green) is constructed at this position in the ultrasound image. Rightmost image: after the graph cut, the segmentation result (red) is displayed to the user. Note: figure adapted from[26].

In this contribution, an interactive real-time segmentation approach for liver metastases in ultrasound images is presented. The purpose of our experiments was to adapt and apply a specific graph-based segmentation scheme (that has been evaluated already on MRI[22] and CT[23] data) to ultrasound data, which is much more challenging and still under active research. A successful automatic segmentation can replace time-consuming manual outlining of pancreatic cancer liver metastases. The usage of our approach is easy and intuitive, and enables a rapid detection of the metastasis and segmentation just by moving the mouse over the image. The interactive real-time segmentation is based on a specific graph-based algorithm[24] that allows a very fast s-t-cut[25] calculation. Furthermore, the proposed algorithm requires only one user-defined seed point to generate the graph. Even if there are semi-automatic methods in the literature that have been used to segment malignant lesions in US acquisitions, the authors are not aware of an interactive real-time segmentation algorithm in this research area. However, note that some initial results of the interactive segmentation have been presented as poster at the SPIE Medical Imaging conference[26].

## Materials and Methods

**Data Acquisition.** Ultrasound acquisitions have been performed by using a multifrequency curved probe. The US probes LOGIQ E9 and Aplio 80 from GE Healthcare (Milwaukee, Il, USA) and Toshiba (Otawara, Japan), respectively, allowed image scans with a bandwidth of one to six MHz. Fulfilling the following criteria, images have been selected retrospectively from the digital picture archive of the Katharinenhospital Stuttgart (Germany) ultrasound unit. Regarding the first part of the study, where the algorithm was adjusted using images with different echo pattern, ultrasound examinations of patients treated for different malignancies were chosen. The main selection criteria consisted of the echogenicity class of the lesion, which could be hypo-, hyper or isoechoic. The selection process was stopped when one lesion per echogenicity class was identified. Additional three images with metastases of different echogenicity were used in the first part of this study to visualize the performance of the algorithm. The selection of 40 images in the second part of the study for the evaluation of the algorithm was done by choosing ultrasound examinations of consecutive patients treated for metastatic pancreatic cancer. However, we excluded images were text or markers overlaid the target lesion. Afterwards, we removed the patient information from the image, and the anonymized image was segmented with our algorithm. The local ethical committee from the Katharinenhospital, Stuttgart, Germany, provided a waiver of the requirement for informed consent for this retrospective study and allowed the publication of anonymized data. Note, we added figure serial numbers to the images in the lower left corners. Cases we used to develop, test and present our algorithm in the first part of the manuscript are indicated with #T followed by a number, e.g. #T1 in in Fig. 2. The 40 cases we used to evaluate our algorithm are indicated with #E followed by a number from 1 to 40, e.g. #E39. The anonymized raw data can be used for own research purposes as long as our work is cited[27]: https://www.researchgate.net/publication/307907688_Ultrasound_Liver_Tumor_Datasets.

**Segmentation process.** Figure 2 outlines the overall high-level workflow of our approach for an interactive detection and segmentation of liver metastases in ultrasound images. The workflow starts in the leftmost image. A circular template is used for the graph construction. The center of the template is a user predefined single point, called the seed point, placed on the image in the center of the area of interest. The placement of the seed point results in the graph construction consisting of multiple rays emitted from the circle's center. In the second image from left, two of these rays are visualized. The graphs' nodes represented by little circles are sampled along the rays, and the different edges between the nodes are set up, whereby a pre-defined delta value $\Delta_r$ influences the inter-edges construction. Additionally, Fig. 3 presents a flow chart of the semi-automatic segmentation approach. For a detailed description regarding the construction of the graph see below under *Algorithm Development*. To illustrate such a graph the third image presents a complete graph overlaid on an US image. Note that, during an interactive segmentation, the complete graph is not shown to the user. Instead, the user sees only the corresponding segmentation result after placement of the seed point. This is presented in the rightmost image where the outline of the segmentation result is marked by red dots and the user-defined seed point is marked by a white dot in the center.

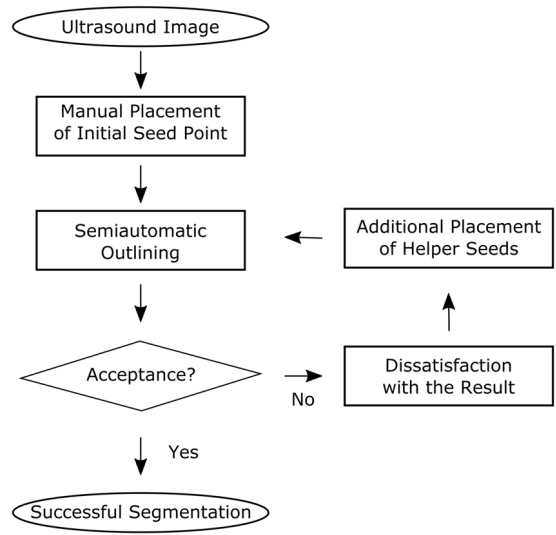

**Figure 3.** Flow chart of the semi-automatic segmentation approach.

**Algorithm Development.** The segmentation approach of this study belongs to the class of graph-based algorithms[28]. Graph-based methods convert an image or parts of it into a graph $G(V, E)$ that consists of nodes $n \in V$ and edges $e \in E$. The nodes $n \in V$ are sampled in the image, except for two virtual nodes $s \in V$ and $t \in V$, which are called the source and the sink, respectively. The two virtual nodes are used to calculate an s-t-cut[25] after the graph construction and divide the graph into two parts: the foreground (e.g., the liver metastasis) and the background (e.g., the surrounding liver tissue). Furthermore, the edges $e \in E$ establish connections between the (virtual) nodes (for example, the edge $\langle v_i, v_j \rangle \in E$ connects the nodes $v_i, v_j$[29]). However, the first step of our approach is to sample the graphs' nodes along radial rays that are equidistantly distributed around a fixed point in a clockwise manner. Figure 4 illustrates this course of action in the upper image) and shows how the so called ∞-weighted intra-edges that connect nodes along the same ray are constructed (lower left image). The ∞-weighted intra-edges ensure that the s-t-cut affects only one of the edges that belong to the same ray, ensuring a star shaped result[30]. Afterwards, the ∞-weighted inter-edges are constructed, as shown in the two rightmost lower images of Fig. 4. The ∞-weighted inter-edges connect nodes from different rays. The delta value $\Delta_r$ influences the number of possible s-t-cuts, hence influencing the smoothness of the segmentation result ($R$ is the number of rays with $r = (0, …, R − 1)$). That means that the greater the delta value, the greater is also the flexibility of the resulting segmentation contour. Figure 5 shows several segmentations of the same metastasis (the user-defined seed point in white remains at the same position for all segmentations). In the leftmost image, the delta value was set to zero (see also Fig. 4 lower image in the middle), thus, the resulting contour has also to follow a predefined template (in this case a circle). The size of the circle, however, depends now purely on the gray values in the image. As one can see, with greater delta values from 1 (second image from the left, see also Fig. 3 lower right image) to 5 (rightmost image), the resulting segmentation contour gets more flexible and is able to adapt to unregular gray value deviations in the image. Note that, for our study, we set the delta value to 2. Next, the weighted edges between the nodes and the source/sink are generated. An average gray value is sampled on the fly around the user-defined seed point. Next, the absolute differences between this average gray value and the gray values behind the sampled nodes are calculated. Finally, the differences between two adjacent absolute values are determined to be the weights of the edges. Figure 6 gives an example for this course of action. The white dot in the middle image is the user-defined seed point. Around this user-defined seed point, an average gray value is calculated (white circle), which is, in our example, around 100. Now we have nodes that are sampled in direction of a brighter surrounding area (blue circle, left values) and nodes that are sampled in direction of a darker surrounding area (green circle, right values). The gray values (GW) give example values that have been sampled in the image. As one can see, these values are around 100 in the tumor area (T) and significant higher and lower in the surrounding areas (S): 131, 160 and 155 in a brighter area (blue circle and example values on the left), and 60, 40 and 55 in a darker area (green circle and example values on the right). The absolute values between the average gray value (100) are shown in the columns $c$ (costs), and the weights for the edges (differences between two adjacent costs) are displayed in the columns $w$. The signs (negative/positive) define if a node is bound to the source or the sink, except for the very first and last nodes, which are bound with their absolute weight values to the source or sink, respectively. As one can see in the example of Fig. 6, there are edges with high costs between the transition from the tumor (T) to the surrounding tissue (S) of 49 and 43, respectively. These edges will not likely be cut by the s-t-cut and, hence, the cut will occur before these edges. In contrast, heterogeneous regions (T: 94, 110, 109 and 95, 101, 98) have similar gray values and no strong edges. Thus, the s-t-cut in these areas depends on the adjacent rays and the delta value $\Delta_r$. In a nutshell, the graph construction replaces the pre-defined templates with fixed seed point positions used in previous works[31, 32], with a circular template centered around a user-defined seed point.

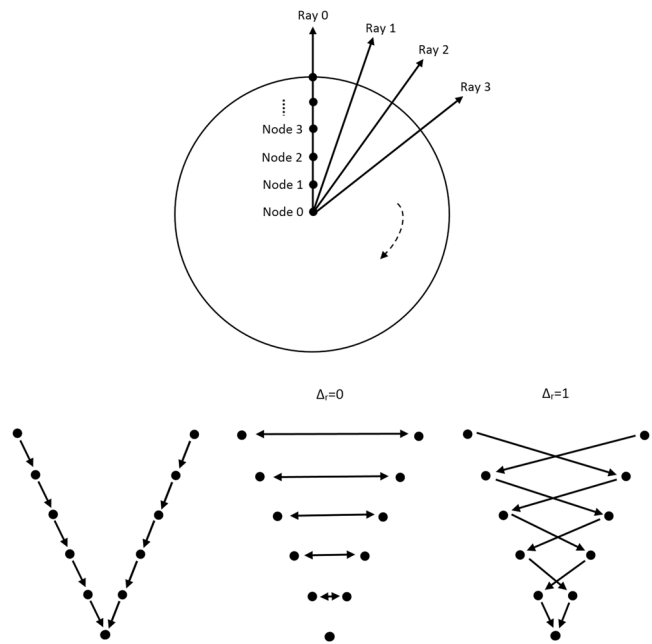

**Figure 4.** Underlying principle of the graph construction for the interactive segmentation: The upper image shows how the nodes for the graph are sampled. Rays are distributed radially clockwise around a fixed point. Along these rays, the nodes (0, 1, …) are sampled. Bottom images: The intra-edges are constructed within the single rays (an example for two rays is shown in the leftmost image). The next step is to construct the inter-edges between the rays under the specification of the delta value $\Delta_r$: for a delta value $\Delta_r$ of zero ($\Delta_r = 0$), only inter-edges are allowed that connect nodes on the same "level" along the rays (see image in the middle). For a delta value $\Delta_r$ of one ($\Delta_r = 1$), inter-edges are allowed that connect nodes on different "levels" along the rays, however, only with a maximum "level" distance of one (see rightmost image). Note: the delta values can also be higher, e.g., 2, 3 and so on. Note: figure adapted from[26].

## Results

### Adjustment of the segmentation algorithm using liver metastases with different echo pattern.
Our segmentation algorithm was integrated into the medical prototyping platform MeVisLab (Fraunhofer MeVis, Bremen, Germany, www.mevislab.de) as a C++ module[33, 34]. The interactive real-time segmentation could be performed smoothly on a Macbook Pro laptop with an Intel Core i7-4850HQ CPU @ 2.30 GHz, 16 GB RAM and Windows 8.1 Professional installed. The adjustment of our segmentation algorithm was done using a set of liver metastases with different echo pattern. For visual inspection, Fig. 7 shows an interactive and a manual segmentation result of a liver metastasis (metastasis of a colon cancer) that appears brighter (hyperechoic), when compared to the surrounding liver tissue. Further, the metastasis has a darker (hypoechoic) halo and exhibits a very low contrast to the surrounding liver parenchyma. Overall, the leftmost image of Fig. 7 presents the native acquisition with a zoomed area of the metastasis. The image in the middle of Fig. 7 shows the manual measurement result of the metastasis (white dotted line between two crosses). Finally, the rightmost image of Fig. 7 presents the interactive segmentation results and the user-defined seed point at this position (red dots and white dot, respectively). In Fig. 8, the interactive segmentation results of several liver metastases (red dots) are presented as follows: Upper Left: a metastasis of a neuroendocrine neoplasm of the pancreas. Upper right: a metastasis of a colon cancer. Lower images: two different views of a metastasis of a uveal melanoma. Compared to the surrounding liver tissue, the metastasis in the upper left image appears brighter (hyperechoic). In contrast, the two metastases in the lower images appear darker (hypoechoic) when compared to the surrounding liver tissue. Finally, the metastasis of the upper right image has bright and dark areas in comparison to the surrounding areas. For all segmentation results of Fig. 8, the white dots represent the corresponding seed points. At these positions, the user was satisfied with the borders of the lesion and stopped the interactive segmentation process by releasing the mouse button.

Figure 9 presents several screenshots from a video, where two metastases of a colon cancer[35] in one image were segmented interactively (from the top to the bottom). The upper image presents the native acquisition. The next image presents the position of the mouse cursor where the user started the interactive segmentation. The third image presents the first segmentation result (red dots) at the position of the mouse cursor. In the following image, the user moved the seed point (white dot) for the segmentation slightly to the right to get a better segmentation.

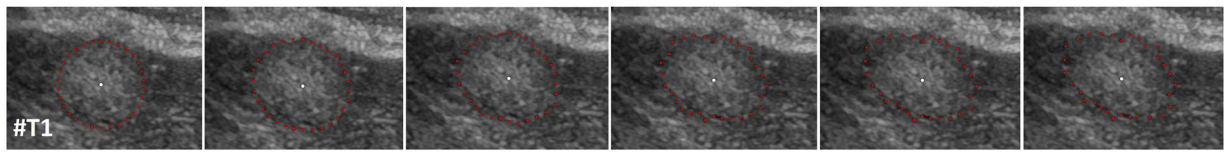

**Figure 5.** Segmentation results for different delta values: $\Delta_r = 0$ (leftmost) to $\Delta_r = 5$ (rightmost). Note, the user-defined seed point in white remains for all segmentations at the same position.

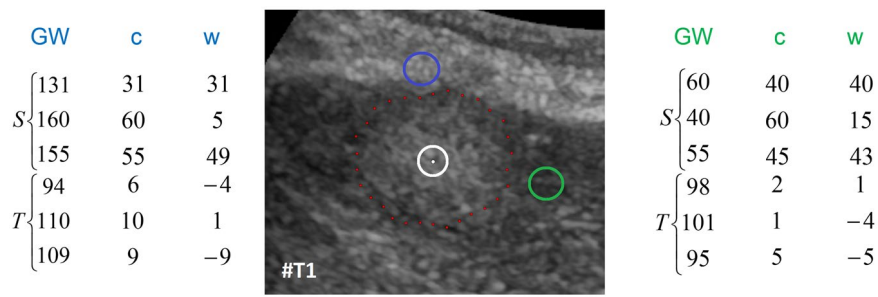

**Figure 6.** Principle of generating the weighted edges between the nodes and the source/sink with an average gray value (for this example 100) sampled around the user-defined seed point (white) in the area of the white circle (US image in the middle). The example on the left belongs to nodes that have been sampled along a ray that runs into a brighter area (blue circle). The example on the right belongs to nodes that have been sampled along a ray that runs into a darker area (green circle). GW are the underlying gray values and the costs $c$ are absolute values between the average gray value (100) and the sampled values behind the nodes (e.g. |100−131| = 31, |100−160| = 60). Finally, the weights $w$ that are assigned to an edge between a sampled node and the source or sink is calculated between two adjacent absolute cost values, e.g. 10−9 = 1. The signs (negative/positive) define if a node is bound to the source or the sink, except for the very first and last nodes (31, 9, 40 and 5), which are bound with their absolute weight values to the source (−9, −5) or sink (31, 40), respectively.

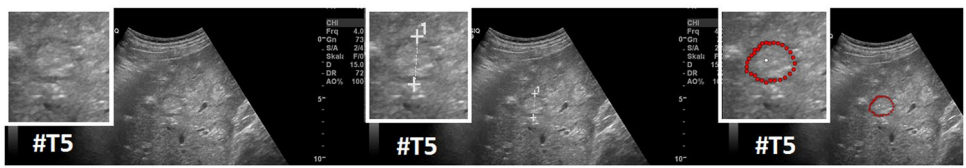

**Figure 7.** Segmentation results (manual/interactive) of a hyperechoic appearing metastasis with a hypoechoic halo (metastasis of a colon cancer), where the metastasis shows a very low contrast to the surrounding liver parenchyma. The native image with a zoomed view of the metastasis is presented in the left image. A manual measurement of the maximal metastasis diameter (white dotted line between two white crosses) is shown in the middle image. Finally, the rightmost image presents the interactive segmentation results (red dots) with the corresponding user-defined seed point (white). Note: figure adapted from[26].

After being satisfied with the segmentation of the first metastasis, the user moved the seed point to the second metastasis on the right. As seen in the screenshots of the video, the resulting segmentation contours (red dots) collapsed, because, in this area, no lesion is present (note: the screenshots present only a fraction of the whole video). When the user reached the second metastasis on the right, the red segmentation contour automatically expanded again and adapted to the metastasis border (lower two images). The lower image presents the final outlining of the second metastasis, where the user stopped the interactive segmentation process. Moreover, Fig. 10 shows a side-by-side comparison of manual expert measurements (left) and interactive segmentation results (right) for the two liver metastases from Fig. 9.

**Evaluation of the segmentation algorithm using pancreatic cancer metastases.** After the initial setup of the algorithm parameters using liver metastases with predefined echo pattern, we evaluated the algorithm using a set of 40 images of 24 patients with pancreatic cancer liver metastases from clinical routine. In total, 37 different metastases have been segmented in random order. Three metastases have been segmented with a different image zoom. The selection reflected the general echogenicity distribution of pancreatic cancer metastases, with 31 of the 37 metastases being hypoechoic, 5 isoechoic and one hyperechoic compared to the surrounding liver tissue. The images have not been previously segmented with the proposed algorithm before. Initially, the operator performed a manual segmentation of the metastases and then measured the maximal diameter in a separate window. Subsequently, the semi-automatic segmentation was performed. Figure 11 presents an overlay of the three measurements. The red outline represents the manual segmentation including the white arrow representing the manually drawn maximum diameter. The yellow line represents the result of the semiautomatic segmentation. The operator spent a total of 26 minutes for the manual measurements of the 40 metastases and 18 minutes using the semiautomatic segmentation. This results in a mean segmentation time of 27 seconds per metastasis using the semi-automatic method and of 39 seconds using the manual method. Comparison of the two methods revealed an average Dice Score (DSC)[36] of $84.76 \pm 5.08\%$ and an average Hausdorff Distance (HD)[37] of $12.58 \pm 5.74$ pixel (Table 1). The DSC is the agreement between two binary volumes and is calculated as follows:

$$DSC = \frac{2 \cdot V(A \cap R)}{V(A) + V(R)} \tag{1}$$

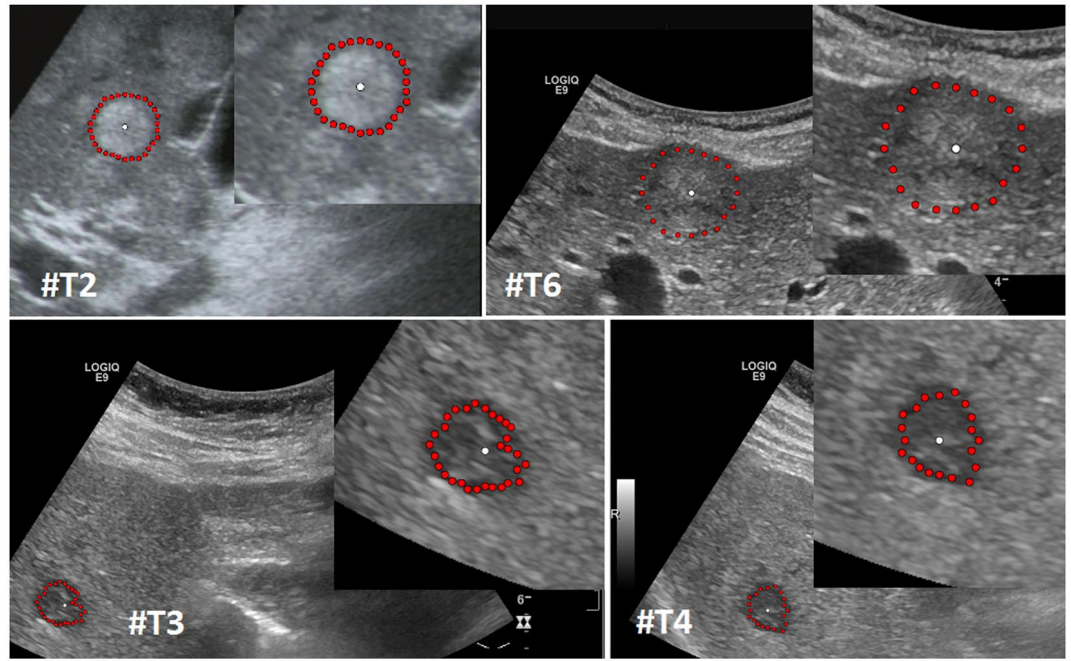

**Figure 8.** Segmentation results for different liver metastases (red dots). Presented are a hyperechoic metastasis of a neuroendocrine neoplasm of the pancreas (upper left), an isoechoic metastasis of a colon cancer (upper right) and two different views of a hypoechoic metastasis of an uveal melanoma (lower images). The white dots in the images are the user-defined seed points. At this positions the user stopped the interactive segmentation, because (s)her was satisfied with the automatic segmentation of the metastasis border. Note: figure adapted from[26].

In summary, the *DSC* measures the relative volume overlap between *A* and *R*, where *A* and *R* are the binary masks from the algorithmic (*A*) and the reference (*R*) segmentation. $V(\cdot)$ is the volume (in mm$^3$) of voxels inside the binary mask, by means of counting the number of voxels, then multiplying with the voxel size. The *HD* between two binary volumes is defined in terms of the *Euclidean* distance between the boundary voxels of the masks. Given the sets *A* (of the algorithmic segmentation) and *R* (of the reference segmentation) that consist of the points that correspond to the centers of segmentation mask boundary voxels in the two images, the directed *HD* $h(A, R)$ is defined as the minimum *Euclidean* distance from any of the points in the first set to the second set, and the *HD* between the two sets $H(A, R)$ is the maximum of these distances[38]:

$$h(A, R) = \max_{a \in A}(d(a, R)), \quad \text{where} \quad d(a, R) = \min_{r \in R} \|a - r\| \tag{2}$$

$$H(A, R) = \max(h(A, R), h(R, A)) \tag{3}$$

Furthermore, we added the number of pixels for the manual and automatic segmentations to Table 1 to have a reference to the size of the tumor (although the reader has to consider that the conversion of pixel to mm depend on the ultrasound probe and the depth of the image). Two examples of challenging images with a DSC of less than 80% are presented in Figs 12 and 13. The manual segmentation outline is represented by the red line, and the semi-automatic segmentation is depicted in yellow. The manually drawn line is much smoother in comparison to the line generated by the algorithm. The algorithm, however, seems to outline the metastasis in a much more precise way. Yet, this results in a poor DSC score due to mismatch of the two areas.

Furthermore, we tested our data with freely available segmentation algorithms called GrowCut[39–41] and Robust Statistics Segmentation (RSS)[42], and looked at the state of the art segmentation methods available under the medical platform MITK (http://docs.mitk.org/2015.05/org_mitk_views_segmentation.html). Aside from convenient options to support manual outlining (like Add, Subtract, Correction, etc.), MITK also offers more advanced 2D and 3D segmentation algorithms for medical structures. However, the interactive (2D) Region Growing and the 2D Fast Marching[43] were not able to segment our cases, because of the small differences in grey values between the liver metastases and the surrounding tissues. The interactive outlining tool called Live Wire[44] is an option, because it snaps to the metastases border during the segmentation process. Although using the tool is cumbersome due to the need of placing each single seed quite close to each other alongside the metastasis contour.

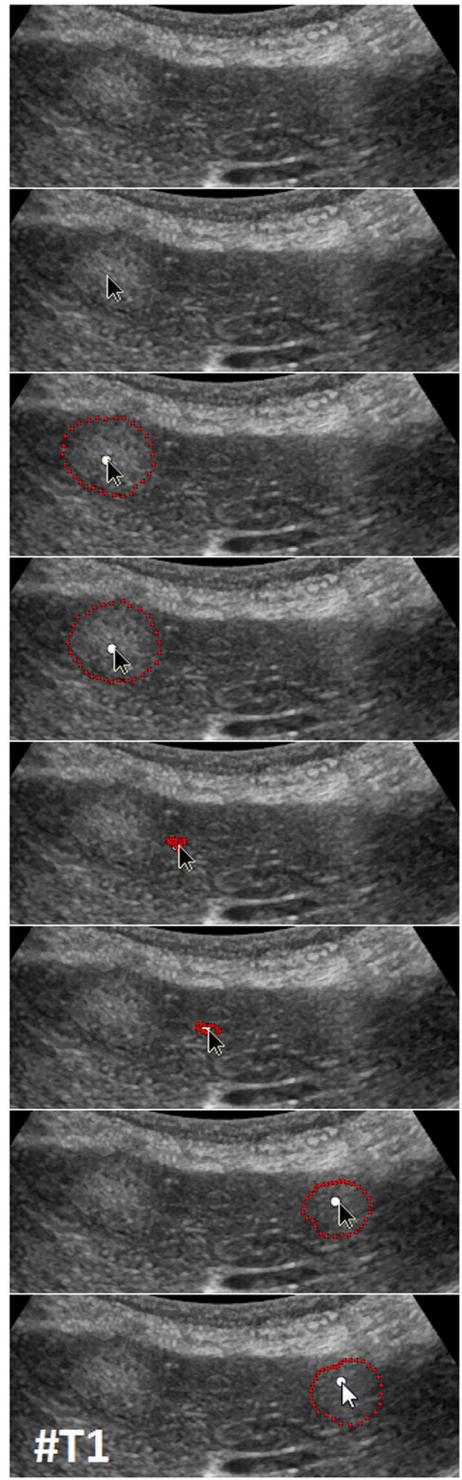

**Figure 9.** Several screenshots from a video where two metastases of a colon cancer in one image were segmented interactively (from the top to the bottom). The upper image presents the native acquisition. The next image presents the position of the mouse cursor where the user started the interactive segmentation. The third image presents the first segmentation result (red dots) at the position of the mouse cursor. In the following image, the user moved the mouse and therefore the seed point (white dot) for the segmentation slightly to the right to get a better segmentation. After being satisfied with the segmentation of the first metastasis, the user moved the mouse to the second metastasis on the right. As seen in the screenshots of the video, the resulting segmentation contours (red dots) collapsed, because in this area, no metastasis is present (note: the screenshots present only a fraction of the whole video). When the user reached the second metastasis on the right, the red segmentation contour automatically expanded again and adapted to the metastasis border (lower two images). The lower image presents the final outlining of the second metastasis where the user stopped the interactive segmentation process.

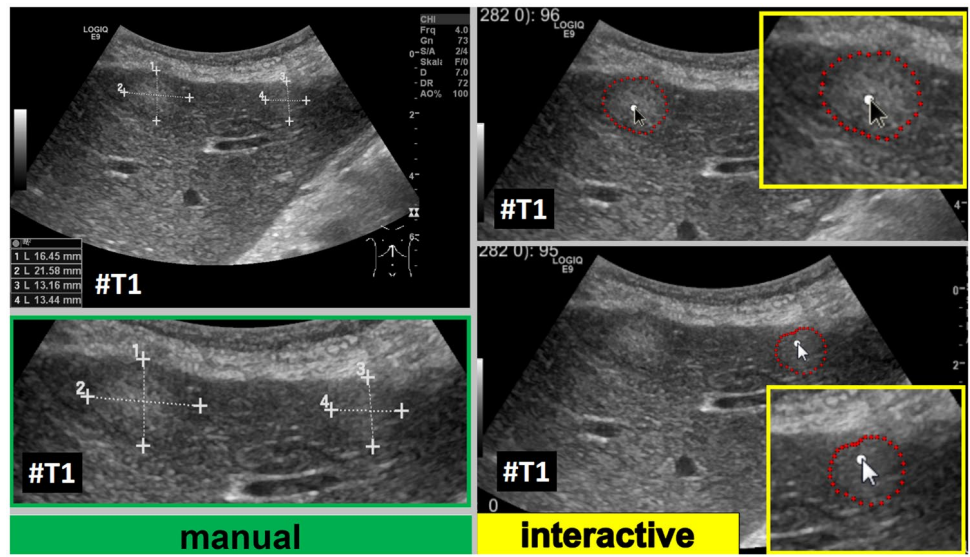

**Figure 10.** Direct side-by-side comparison of the interactive segmentation results for the two colon cancer liver metastases from Fig. 9 (right side) and a manual expert measurement of the metastasis (left side). Note: figure adapted from[26].

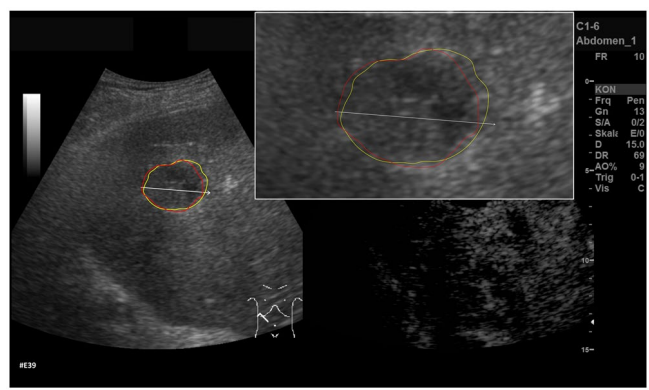

**Figure 11.** Comparison of manual and semi-automated segmentation of a pancreatic cancer liver metastasis. The native image with a zoomed view of the metastasis is presented (white box). The red outline represents the manual segmentation including the white arrow representing the manually drawn maximum diameter. The yellow line represents the result of the semi-automatic segmentation.

Figure 14 shows example segmentation results of GrowCut for the two metastases from Fig. 9. The left images of Fig. 14 show the manual initialization of GrowCut. The metastases were initialized with green and the backgrounds were initialized with yellow (note: an equivalent initialization has been used for the RSS). The images in the middle present the segmentation results of GrowCut (green) for both metastases. Finally, the two rightmost images show a closer view of the segmentation results (green) with a lower opacity to see the structures of the US image behind the segmentation masks. Analogous to Figs 14 and 15 shows a GrowCut segmentation result for the pancreas metastasis from Fig. 11. Again, the left image shows the manual initialization of GrowCut: The pancreas metastasis was initialized with green, and the background was initialized with yellow. The image in the middle shows the segmentation result of GrowCut (green). The right image shows a closer view of the segmentation result (green) with a lower opacity of 0.6 to see the structures of the US image behind the segmentation mask. Figure 16 presents a direct comparison of our approach with GrowCut (green) for case 39. The red outline represents the manual segmentation including the white arrow representing the manually drawn maximum diameter. The yellow line represents the result of the semi-automatic segmentation (see also Fig. 10). Compared with the manual segmentation, our semi-automatic segmentation algorithm could achieve a DSC of 92.47%. However, compared with the manual segmentation, the GrowCut-based segmentation achieved a DSC of 85.46%. Finally, Table 2 presents a direct comparison for the four cases where our approach could achieve a DSC higher than 90% and the six cases where our approach achieved a DSC lower than 80%. For the RSS, several parameters had to be defined: The Approximate Volume, the Intensity Homogeneity, the Boundary Smoothness and the Max running time (min).

| | Maximal Diameter (mm) | | Area (mm²) | | Pixel | | DSC (%) | HD (pixel) | HS |
|---|---|---|---|---|---|---|---|---|---|
| | US-Cut | Reader | US-Cut | Reader | US-Cut | Reader | | | |
| Min. | 9.05 | 8.61 | 40.77 | 32.38 | 331 | 370 | 71.62 | 4.12 | 0 |
| Max. | 40.08 | 36.33 | 1038.29 | 777.38 | 17919 | 23916 | 93.62 | 29,15 | 8 |
| µ±σ | 20.17±7.86 | 18.69±6.48 | 287.59±229.04 | 232.09±191 | 3709.96±4439.3 | 4905.44±5621.73 | 84.76±5.08 | 12.58±5.74 | 4.33±1.95 |

**Table 1.** Manual vs algorithmic (US-Cut) results for 40 cases of pancreatic cancer hepatic metastases: Dice Similarity Coefficient (DSC), Hausdorff Distance (HD), helper seeds (HS)[58], mean value (µ) and standard deviation (σ).

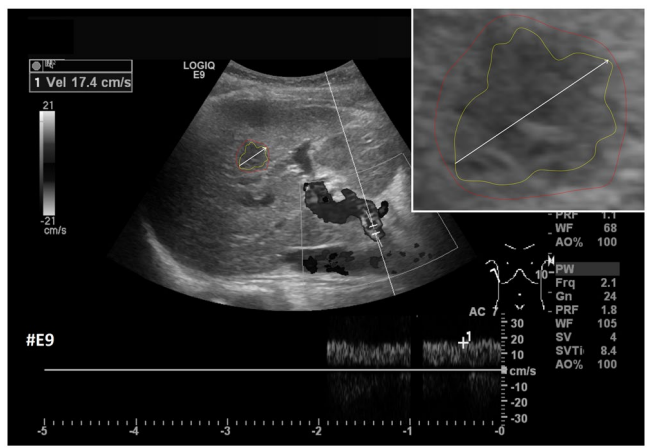

**Figure 12.** Example of a segmentation with a DSC of less than 80%. Depicted is the native image with a zoomed view of the metastasis (white box). The red outline represent the manual segmentation including the white arrow representing the manually drawn maximum diameter. The yellow outline represents the result of the semi-automatic segmentation.

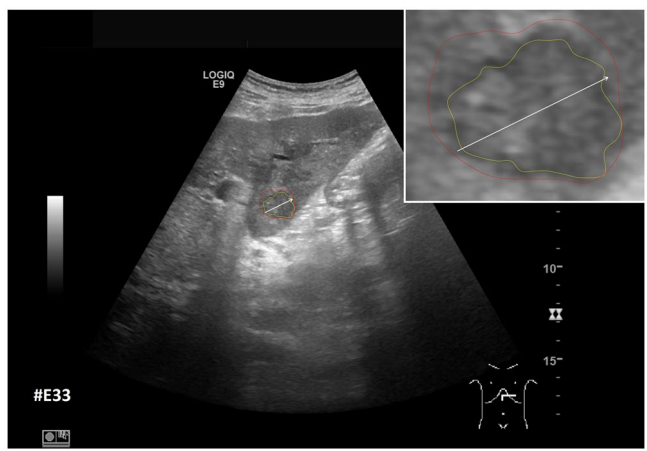

**Figure 13.** Example of a segmentation with a DSC of less than 80%. Depicted is the native image with a zoomed view of the metastasis (white box). The red outline represent the manual segmentation including the white arrow representing the manually drawn maximum diameter. The yellow outline represents the result of the semi-automatic segmentation.

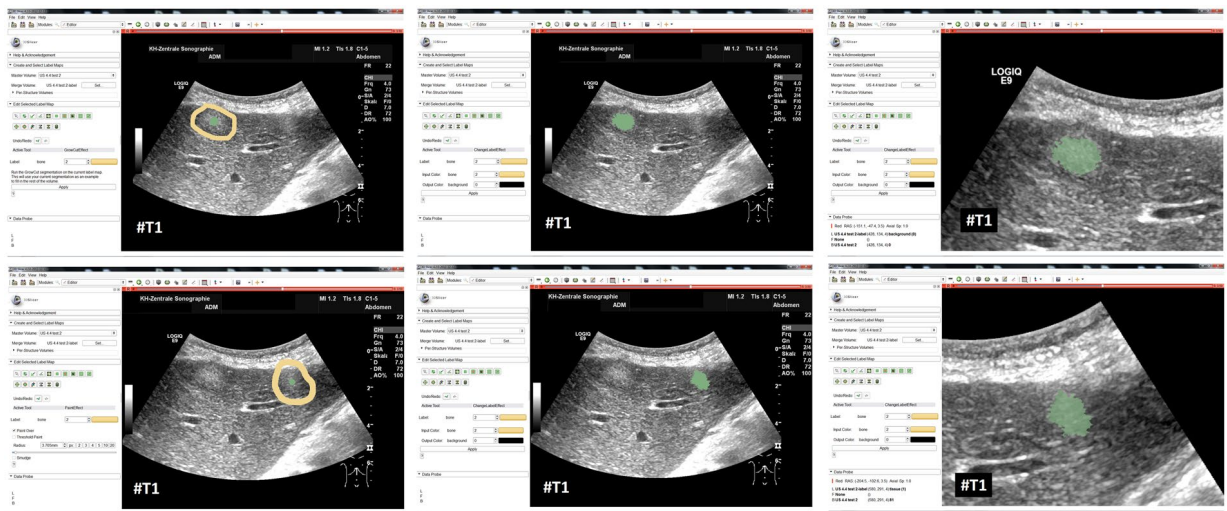

**Figure 14.** GrowCut segmentation results for the two metastases from Fig. 9. The left images show the manual initialization of GrowCut: the metastases were initialized with green, and the backgrounds were initialized with yellow. The images in the middle show the segmentation results of GrowCut (green). The right images show a closer view of the segmentation results (green) with a lower opacity.

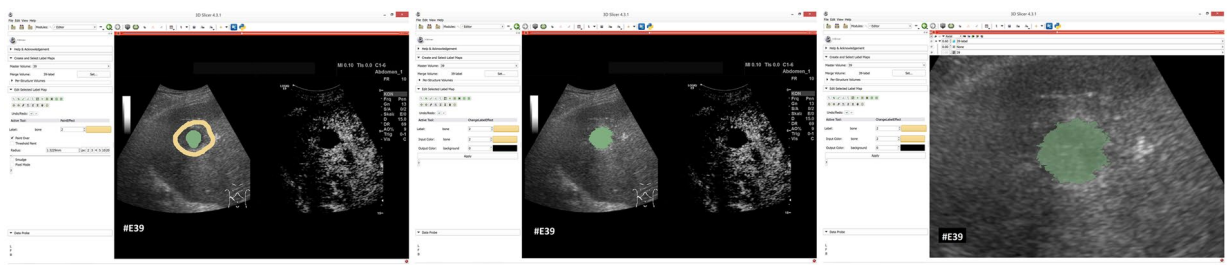

**Figure 15.** GrowCut segmentation results for a pancreas metastasis. Equivalent to Fig. 14, the left image shows the manual initialization of GrowCut: the pancreas metastasis was initialized with green, and the background was initialized with yellow. The image in the middle shows the segmentation result of GrowCut (green). The right image shows a closer view of the segmentation result (green) with a lower opacity (0.6).

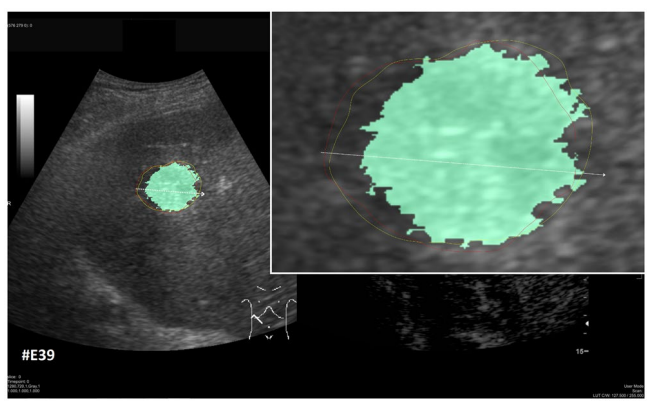

**Figure 16.** Direct comparison of our approach with GrowCut (green) for case 39. The red outline represents the manual segmentation including the white arrow representing the manually drawn maximum diameter. The yellow line represents the result of the semi-automatic segmentation (see also Fig. 11). Compared with the manual segmentation, our semi-automatic segmentation algorithm could achieve a DSC of 92.47%. However, compared with the manual segmentation, the GrowCut-based segmentation achieved a DSC of 85.46%.

Firstly, we identified a parameter setting that achieved a high DSC for the first case in Table 2 (case 4). For an Approximate Volume of 40, an Intensity Homogeneity of 0.6, a Boundary Smoothness of 0.6 and a Max running time (min) of 10.0, we were able to achieve a metastasis segmentation result yielding a DSC of over ninety percent (90.55%). Afterwards, we used these parameter settings for the remaining cases with mixed results, for example leaking into surrounding tissues resulted in low DSCs for cases 9, 17, 28, 33 using RSS. However, we could achieve an acceptable DSC of 84.01% using RSS for segmenting the difficult case 8, which had the lowest score for our approach (DSC 71.62%) and GrowCut (DSC 65.2%). This was only possible after trying different parameter combinations for Intensity Homogeneity, Boundary Smoothness and the Approximate Volume. Finally, for case 39 we could not achieve any reasonable result, because it always leaked massively into the surrounding tissue, despite different parameter settings.

| Case | DSC (%) | | |
|---|---|---|---|
| | US-Cut | GrowCut | RSS |
| 4 | 92.11 | 91.84 | 90.55 |
| 6 | 93.62 | 86.23 | 86 |
| 8 | 71.62 | 65.2 | 58.53 |
| 9 | 76 | 86.8 | 50.6 |
| 14 | 77.04 | 84.6 | 74.06 |
| 17 | 77.28 | 86.4 | 53.62 |
| 23 | 90.73 | 82.23 | 75.78 |
| 28 | 74.59 | 69.19 | 59.99 |
| 33 | 78.34 | 88.77 | 53.56 |
| 39 | 92.47 | 85.46 | — |

**Table 2.** Direct comparison for the four cases where our approach could achieve a DSC higher than 90% and the six cases where our approach achieved a DSC lower than 80%.

## Conclusions

In this contribution, we presented the results of a fast interactive segmentation algorithm for liver metastases in ultrasound acquisitions, which is a challenging problem because of the low quality of the image data and the low contrast between metastases and surrounding tissue. As a result, liver metastases are still measured in a purely manual way in the clinical routine. This leads to a poor inter-observer agreement regarding the size of the metastasis, which is the critical factor for treatment response evaluation and treatment planning. Moreover, the metastasis volume is only approximated via diameters instead of outlining the whole metastasis border. Our proposed method supports manual metastasis outlining by applying an interactive real-time segmentation algorithm to the ultrasound images, which gives the user an instant feedback of the metastasis border. The segmentation algorithm was developed and tested hand-in-hand by computer scientists and physicians to ensure a practical usage in a clinical setting. The first step included an initial setup of segmentation parameters using metastases with different echogenic properties including hyperechoic, hypoechoic or isoechoic metastases. Afterwards, the optimized parameters were used to evaluate the algorithm using a set of 40 pancreatic cancer liver metastases images in comparison to a manual assessment of the metastases size. The statistical evaluation of the algorithm covered standard acquisitions from the clinical routine. Especially the interactive real-time behavior of the approach allowed satisfying segmentation result within a few seconds without parameter tweaking. This stands in strong contrast to fully-automatic approaches that still fail too often and, therefore, are not accepted by the users for daily usage. Our method stands also in contrast to semi-automatic approaches, like GrowCut, which need some kind of initialization, like marking parts of the lesion and the surrounding tissues, before they can be executed. In general, these approaches can achieve good results because of the additional information about the location and the appearance of the pathology, but also need time to reinitiate the segmentation process in case the segmentation fails.

The research highlights of our work consist of:

- The development of a special interactive segmentation approach;
- Interactive real-time segmentation feedback for the user;
- Intuitive and fast achievement of satisfying segmentations within seconds;
- Statistical evaluation with acquisitions from the clinical routine.

Furthermore, we tested our data with the GrowCut implementation that is available with 3D Slicer ([www.slicer.org](www.slicer.org)). In recent studies, the algorithm showed that it can support the time-consuming manual volumetry in medical data, like lung cancer[40], glioblastoma multiforme (GBM)[38] and pituitary adenoma[41]. We also found that, for a careful initialization, GrowCut can return reasonable segmentation results for our US data. However, in contrast to 3D, the time and effort for the 2D initialization on our data is about the same as segmenting the metastasis manually (not taking into account the executing time for GrowCut to calculate the segmentation result after the initialization, which is hardware depended). However, due to the noisy US images without a clear border between the tumor and the surrounding structures, the segmentation results looks *frazzled*. Hence, without further image processing, it is hard to determine the maximal diameters within the segmentation result/mask (unlike to GBMs where you have a clear border in the contrast-enhanced MRI images). Nevertheless, GrowCut can make more sense for more complex 2D structures, where manually outlining takes some time, and GrowCut can be initialized with some rough strokes.

In contrast to existing methods[5–11], our approach focuses on a very fast interactive segmentation that can be performed in *real-time*. Hence, our approach is also eligible for interventions, where images are acquired intra-operatively, such as during liver tumor ablations, where the ablation needle is image-guided via intraoperative US and CT scans. Here, our approach enables an instant metastasis segmentation with a subsequent registration to the high-resolution pre-operative planning acquisition. In comparison to Bui *et al.*[45], we could achieve a similar Dice score, but with real rather than simulated data. A limitation of our study is that we only had one manual segmentation and even different experts (radiologists) would make somewhat different delineations. In previous work[46, 47], we studied the inter-observer Dice scores for manual delineations of ablations of liver tumors in CT acquisitions, which yielded inter-observer DSCs of 86.8% and 88.8%, respectively. This shows how close we are with our 85% DSC to pure manual segmentations. Note that CT segmentations are, in general, easier to achieve than US because of the high contrast between tumor border and surrounding tissues. Further, we studied the segmentation results of GBMs from several surgeons (one surgeon even segmented the cases twice to evaluate also the intra-surgeon deviation) and showed the variations[48]. In contrast to some state-of-the-art methods, our algorithm can only handle star-shaped structures[30], which limit its applicability (note that for our scenario and all studied cases this was sufficient). However, a first solution to this limitation has been proposed by Baxter *et al.*[49].

There are several areas for future work, like the monitoring of the same liver lesions over time. Additionally, the correlation of segmentation results assessed by different clinical examiners will provide more insight in the usability and accuracy of the algorithm in clinical routine. The algorithm needs to be tested on additional sub-groups of tumor origin like colorectal or breast cancer. The long term goal is to ease the registration of liver masses in an everyday practice. Additionally, this will help to fuse intraoperative US images via the interactive segmentations to the patient's pre-interventional computed tomography acquisitions. Thus, supporting local ablative interventions of liver tumors by guiding a precise needle placement for tumor ablations, like for radiofrequency ablation (RFA) in the European funded ClinicIMPPACT project[46, 47]. Furthermore, we started to acquire and collect 3D US datasets of liver metastases for a 3D (volumetric) tumor analysis, which is even more time-consuming when performed manually[50]. For a 3D segmentation, we will apply a spherical[51–53] instead of a circular template for the interactive real-time segmentation. In the long run, we plan to further examine these lesions using a virtual reality environment accessed via a head-mounted display (HTC Vive) integrated in MeVisLab[54] and a system based on augmented reality using an optical see-through head-mounted display[55].

Finally, besides the tumor volume, there are other aspects, like the tumor shape, that can be considered for evaluating the liver metastases. These aspects can be ramifications, lobulations or margins. However, most of the hepatic metastases derived from gastrointestinal tumors or primary hepatic tumors have a spherical shape due to clonal expansion of the tumor cells and the homogenous environment they grow in – unlike primary tumors that, for example, grow alongside vascular structures. Thus, our algorithm did not focus on ramification of measurement of lobe development in this study. Despite this, once the tumors are segmented, the tumor masks can relatively easily be analyzed regarding their shapes. This can, for example, be achieved by fitting an ellipse to the segmented tumor mask to see how *roundish* or *longish* the liver metastases appear[56, 57]. Afterwards, calculating the ratio between the two maximal diameters that are orthogonal to each other, works automatically.

### Acknowledgements
The work received funding from BioTechMed-Graz ("Hardware accelerated intelligent medical imaging"), the 6th Call of the Initial Funding Program from the Research & Technology House (F&T-Haus) at the Graz University of Technology (PI: Dr. Jan Egger) and was supported by the TU Graz Open Access Publishing Fund. Dr. Xiaojun Chen received support from Natural Science Foundation of China (Grant No.: 81511130089), Foundation of Science and Technology Commission of Shanghai Municipality (Grant No.: 14441901002, 15510722200, and 16441908400). A video demonstrating the interactive real-time segmentation of liver tumors in an Ultrasound image can be found under the following YouTube-channel: https://www.youtube.com/c/JanEgger/.

### Author Contributions

Conceived and designed the experiments: J.E. and A.H. Performed the experiments: J.E. and A.H. Analyzed the data: A.H. and J.E. Contributed reagents/materials/analysis tools: J.E., D.S., X.C., W.Z. and A.H. Wrote the paper: J.E. and A.H.

### Additional Information

**Competing Interests:** The authors declare that they have no competing interests.

**Publisher's note:** Springer Nature remains neutral with regard to jurisdictional claims in published maps and institutional affiliations.